\documentclass[letterpaper]{article} 
\usepackage{aaai25}  
\usepackage{times}  
\usepackage{helvet}  
\usepackage{courier}  
\usepackage[hyphens]{url}  
\usepackage{graphicx} 
\usepackage{natbib}  
\usepackage{caption} 
\usepackage{algorithm}
\usepackage{algorithmic}
\usepackage{amsmath}
\usepackage{amssymb}
\usepackage{amsfonts}
\usepackage{enumitem}
\usepackage{booktabs}
\usepackage{subcaption}
\usepackage{comment}
\usepackage{tabularx}

\newcommand{\textdist}{\mathrm{dist}}
\newcommand{\Obj}{\mathrm{Obj}}

\usepackage{newfloat}
\usepackage{listings}
\DeclareCaptionStyle{ruled}{labelfont=normalfont,labelsep=colon,strut=off} 
\floatstyle{ruled}
\newfloat{listing}{tb}{lst}{}
\floatname{listing}{Listing}
\nocopyright

\title{A Database-Driven Framework for 3D Level Generation with LLMs}
\author{
    Kaijie Xu, Clark Verbrugge\\
}
\affiliations{
    Department of Computer Science, McGill University\\
    Montreal, Quebec, Canada\\
    kaijie.xu2@mail.mcgill.ca, clump@cs.mcgill.ca
}

\begin{document}

\maketitle

\begin{abstract}
Procedural Content Generation for 3D game levels faces challenges in balancing spatial coherence, navigational functionality, and adaptable gameplay progression across multi-floor environments. This paper introduces a novel framework for generating such levels, centered on the offline, LLM-assisted construction of reusable databases for architectural components (facilities and room templates) and gameplay mechanic elements. Our multi-phase pipeline assembles levels by: (1) selecting and arranging instances from the Room Database to form a multi-floor global structure with an inherent topological order; (2) optimizing the internal layout of facilities for each room based on predefined constraints from the Facility Database; and (3) integrating progression-based gameplay mechanics by placing components from a Mechanics Database according to their topological and spatial rules. A subsequent two-phase repair system ensures navigability. This approach combines modular, database-driven design with constraint-based optimization, allowing for systematic control over level structure and the adaptable pacing of gameplay elements. Initial experiments validate the framework's ability in generating diverse, navigable 3D environments and its capability to simulate distinct gameplay pacing strategies through simple parameterization. This research advances PCG by presenting a scalable, database-centric foundation for the automated generation of complex 3D levels with configurable gameplay progression.
\end{abstract}

\section{Introduction}
\label{sec:introduction }

Procedural Content Generation (PCG) is important for automating game environment creation, enabling scalable production of diverse assets while supporting mixed-initiative, designer-guided control \cite{Togelius2011,Shaker2016}. Early work focused on 2D layouts and grid‐based puzzles, but modern demands for 3D spaces with complex hierarchies pose new challenges: vertical navigation, reasonable pathfinding, and alignment of spatial layouts with gameplay progression \cite{Summerville2018,Liu2021,Khalifa2020}.

Existing approaches address these challenges in isolation. Search-based methods \cite{Togelius2011} explore vast design spaces but often sacrifice logical coherence. Machine learning techniques \cite{Summerville2018} replicate patterns from existing levels but struggle with efficiency and adaptability. Optimization-based adaptations \cite{10645627} improve spatial arrangements but neglect dynamic gameplay constraints like puzzle sequencing. While Large Language Models (LLMs) and constraint-based optimization have shown promise for detailed level layout within individual spaces \cite{cons2025}, achieving scalable and modular generation of expansive, interconnected multi-floor environments with integrated and controllable gameplay progression remains a significant challenge. 

To address these gaps, we combine database–centric content definition with constraint-driven optimization. An LLM populates three reusable databases—facilities, room templates, and mechanic components—capturing geometry, semantics, and placement rules. At generation time, the system (i) assembles a multi-floor layout by selecting and tiling room templates under inter-room constraints, (ii) optimizes each room’s internal facility layout under local constraints, (iii) inserts mechanic components according to the topological order, and (iv) applies a two-stage repair pass to guarantee navigability. This modular architecture allows designers large-scale structural control and fine-tuned pacing adjustments via simple parameter tweaks. By shifting from specifically designed, real-time LLM prompts to \textbf{an offline, thematic, controllable, and reusable asset repository}, we empower designers to efficiently assemble constraint-satisfied 3D levels: they select assets, adjust a handful of parameters, and obtain a playable environment. Each database remains \textbf{independently extensible}—new room types, constraints, or puzzle mechanics integrate smoothly—supporting rapid iteration and precise control over both global structure and gameplay flow.
Our primary contributions are:
\begin{itemize}
    \item LLM-assisted offline creation of reusable databases (facilities, rooms, mechanics) for scalable, controllable 3D multi-floor level generation.
    \item A hierarchical pipeline combining database-driven room assembly with constraint-optimized facility layouts.
    \item A configurable, topology-aware method for integrating gameplay mechanics to achieve adaptable pacing.
    \item A two-phase repair system for navigational integrity.
    \item Experimental validation of simulating diverse gameplay pacing strategies via database parameterization.
\end{itemize}

\section{Background}
\label{sec:background}

Procedural Content Generation (PCG) automates the creation of game assets, reducing direct human input. Early PCG methods employed rule-based and search-based techniques, enhancing replayability and efficiency but often lacking consistency and quality control \cite{Togelius2011,Shaker2016}. More recently, machine learning and reinforcement learning allowed for adaptive content generation responsive to players and environments \cite{Summerville2018,Liu2021,Khalifa2020}. Facility Layout Problem (FLP)—traditionally focused on optimizing spatial arrangements based on operational criteria \cite{AHMADI2017158,SinghSharma2006}—has been adapted for architectural and game-level design, ensuring compliance with spatial constraints \cite{Martin2006Procedural,Lopes2010ConstrainedGrowth,10645627}. However, transitioning from 2D grid-based levels to complex 3D environments introduced challenges in managing spatial hierarchies and multi-floor connectivity.
Constraint-based models try to address these challenges \cite{smith2011answer, neufeld2015procedural, glorian2021dungeon, cooper2022sturgeon, cooper2023sturgeon, cooper2024literally}, but still lack flexibility. They often require manual encoding of high-level design intent into low-level rules, which we automate by using LLMs to quantify designer concepts into concrete constraints.

Recently, LLMs have significantly impacted PCG. Applications include synthesizing code for modeling software \cite{hu2024}, generating room layouts and meshes from textual constraints \cite{fang2023ctrl,schult24controlroom3d}, and creating interactive 3D scenes from casual sketches \cite{xu2024}. Other research explores LLM-driven narrative and world generation \cite{nasir2024word2world}, controlled parameter definition for level generators \cite{jiang2022}, and direct generation of game levels \cite{todd2023level,nasir2023practical}. Additionally, some works integrate semantic understanding from LLMs with diffusion models to enhance scene realism and detail \cite{rio2024,10550742}.

Despite their progress, real-time LLM pipelines still suffer from unpredictable variability, limited designer control, and high maintenance costs. Prior work \cite{cons2025} mitigated these problems by using LLMs only to extract single-room constraints, leaving generation to classic algorithms, but it cannot scale to multi-floor layouts or model gameplay progression. We therefore propose a fully offline, database-driven pipeline: the LLM seeds a reusable library once, after which optimization algorithms assemble levels. Eliminating live LLM calls saves compute, improves coherence and predictability, and gives designers finer control and easier upkeep than prompt-driven approaches.

\section{Method}
\label{sec:method}
\newcommand{\mathbbpone}[1]{\mathbb{1}_{#1}}

We formalize prior LLM‐driven, single‐room generators with a five‐phase pipeline for automated, multi‐floor 3D level creation. First, LLMs construct three offline databases, each encoding geometry, tags, and placement constraints. Second, we place room templates into a stacked floor plan, recording a global topological order. Third, each room’s contents are arranged via constrained optimization to satisfy collision, proximity, visibility, and orientation rules. Fourth, progression mechanics are placed in appropriate rooms, enforcing a reasonable game flow. Finally, a two‐stage repair layer guarantees complete navigability and playability.

\subsection{LLM-Driven Construction of Composable Facility, Room, and Gameplay Mechanics Databases}
\label{sec:databases}

The foundation of our framework is the offline construction of three LLM-enhanced, reusable databases covering: (1) facilities (objects and entities), (2) room templates (thematic layouts), and (3) gameplay mechanic components. This modular, database-driven approach allows for systematic control over content diversity, and provides a foundation for assembling complex, multi-floor 3D levels. Rather than generating each level from scratch, we reuse these predefined modules to guarantee consistency and logical coherence while minimizing composition effort. Table~\ref{tab:database_constraint_specifications} summarizes the constraint types and key properties.

\begin{table*}[t]
\centering
\resizebox{\textwidth}{!}{
\begin{tabular}{|l|l|p{5.5cm}|p{6cm}|}
\hline
\textbf{Constraint/Property Type} & \textbf{Format Example} & \textbf{Penalty/Logic Example} & \textbf{Description / Purpose} \\ \hline
\multicolumn{4}{|c|}{\textbf{Facility-Level (Intra-Room Relationships and Local Placement; Part of Facility Database)}} \\ \hline
Positional (Custom Axis) & User-defined function \(f(x_c, y_c, W, L) \rightarrow \delta_{\text{val}}\) or \(f(z_c, H) \rightarrow \delta_{\text{val}}\) & \(\text{penalty} = w_{\text{axis}} \cdot (\delta_{\text{val}})^2\) & Enforces custom positional rules along axes (e.g., keep facility centered, on floor). \\ \hline
PlaceInRange & PlaceInRange(F, P1, P2) & Penalty if F is outside box [P1, P2]. & Constrains F to lie within a bounding box. \\ \hline
PlaceByWall & PlaceByWall(F, O) & \(\text{penalty} = w_{\text{wall}} \cdot (\textdist(F,\text{wall}) + |\theta_F - O|)^2\). & Requires F near a wall, optionally with orientation O. \\ \hline
Near (Object-Object) & Near(F1, F2) & For \(d_{12} > d_{\min}\), \(\text{penalty} = w_{\text{near}} \cdot (d_{\min} - d_{12})^2\). & Requires F1 to be close to F2. \\ \hline
Far (Object-Object) & Far(F1, F2) & For \(d_{12} < d_{\max}\), \(\text{penalty} = w_{\text{far}} \cdot (d_{12} - d_{\max})^2\). & Requires facility F1 to be distant from F2. \\ \hline
CanSee & CanSee(F1, F2) & If obstructed, \(\text{penalty} = w_{\text{can\_see}} \cdot \mathbbpone{obs}\). & Ensures a clear line of sight between F1 and F2. \\ \hline
Focus & Focus(F, Target) & If \(\phi > \phi_{\text{th}}\), \(\text{penalty} = w_{\text{focus}} \cdot (\phi - \phi_{\text{th}})^2\). & Orients facility F to face a target (facility or point). \\ \hline
Alignment & Alignment(F1, F2, Axis) & \(\text{penalty} = w_{\text{align}} \cdot \theta^2\). & Aligns F1 with F2 along a specified axis. \\ \hline
Orientation & Orientation(F1, F2) & \(\text{penalty} = w_{\text{orient}} \cdot \theta^2\). & Matches the orientation of F1 with F2. \\ \hline
\multicolumn{4}{|c|}{\textbf{Room-Level (Inter-Room Relationships and Global Placement; Part of Room Database)}} \\ \hline
Positional Preference (Custom Axis) & User-defined function \(f(x_c, y_c, W, L) \rightarrow \delta_{\text{val}}\) or \(f(z_c, H) \rightarrow \delta_{\text{val}}\) & \(\text{penalty} = w_{\text{pos\_room}} \cdot (\delta_{\text{val}})^2\) & Guides placement of room R to a general area (e.g., ``Lobby'' near level entrance). \\ \hline
Adjacency (Room-Room) & AdjacentTo(R1, TargetR) & \(\text{penalty} = w_{\text{adj}} \cdot \min(\textdist(\text{R1}, \text{R}_{\text{target}}))\) if no adj. target. & Stipulates room R1 should be next to a room of type TargetR. \\ \hline
Separation (Room-Room) & SeparateFrom(R1, TargetR) & \(\text{penalty} = w_{\text{sep}} / (\min(\textdist(\text{R1}, \text{R}_{\text{target}})) + \epsilon)\) & Stipulates room R1 should not be close/adjacent to TargetR. \\ \hline
Frequency Control & MaxInstances(R, Count) & Enforced during generation. & Limits how many times room R can appear. \\ \hline
Architectural Type & SetType(R, TypeString) & Definitional. & Defines room R as ``enclosed'' or ``open,'' influencing connectivity. \\ \hline
\multicolumn{4}{|c|}{\textbf{Gameplay Mechanic Component-Level (Progression Logic; Part of Mechanics Database)}} \\ \hline
Standard Facility Constraints & StandardFacilityConstraint($m_i$, $r_i$) & (Applied as per Facility-Level) & Mechanic components can possess any standard spatial/contextual constraints applicable to facilities (e.g., a `Locker Key'' Near a Desk''). \\ \hline
Topological Precedence & Precedes($M_i$, $M_j$) & $\text{penalty} = w_{1}\bigl[\max\bigl(0,\tau(R_i)-\tau(R_j)\bigr)\bigr]^2$ & Enforces that $M_i$’s room precedes $m_j$’s in topological order: $\tau(R_i)<\tau(r_j)$. \\ \hline
Topological Near & TopologicalNear($M_i$, $M_j$, $d^{\max}_{ij}$) & $\text{penalty} = w_{3}\bigl[\max\bigl(0,|\tau(R_i)-\tau(R_j)|-d^{\max}_{ij}\bigr)\bigr]^2$ & Encourages $M_i,M_j$ to be within $d^{\max}_{ij}$ steps in the topological sequence. \\ \hline
Topological Far & TopologicalFar($M_i$, $M_j$, $d^{\min}_{ij}$) & $\text{penalty} = w_{4}\bigl[\max\bigl(0,d^{\min}_{ij}-|\tau(R_i)-\tau(R_j)|\bigr)\bigr]^2$ & Ensures $M_i,M_j$ are separated by at least $d^{\min}_{ij}$ steps topologically. \\ \hline
\end{tabular}
} 
\caption{Constraint and Property Specifications for Facility, Room, and Gameplay Mechanic Databases. Penalty formulas illustrate how deviations from constraints might be quantified during optimization, or how rules are enforced. Variables such as \(w_{\text{type}}\) represent weighting factors; \(d\), \(d_{12}\), \(d^{\min}\), \(d^{\max}\) distances; \(\theta\) angular differences; \(\phi\) view angles; \(\delta_{\text{val}}\) a deviation value from a custom function; \(\mathbbpone{obs}\) an indicator function for obstruction; \(\tau(X)\) the topological order of X. F, F1, F2 denote facilities; P, P1, P2 points; O orientation; R, R1, R2, \(\text{R}_{\text{target}}\), \(\text{R}_{\text{i}}\), \(\text{R}_{\text{j}}\) room types/instances; M, M1, M2, \(\text{M}_{\text{i}}\), \(\text{M}_{\text{j}}\) mechanic components.}

\label{tab:database_constraint_specifications}
\end{table*}

\subsubsection{The Facility Database}
\label{sec:facility_database}

The Facility Database includes all basic units that can populate the game world, ranging from static architectural elements like ``IV stands'' to interactive items and unsettling entities such as ``Nurse Enemies,'' all specialized to our survival horror ``Silent Hill (SH) Hospital'' theme. Each entry in this database contains key properties for layout optimization and semantic understanding.

Key properties of a facility entry include:
\begin{itemize}
    \item \textbf{Name:} A unique identifier (e.g., ``Operating Table'').
    \item \textbf{Dimensions:} A definition of its physical bounding box.
    \item \textbf{Positioning Category:} A designation indicating if its position is fixed or adaptable (static or variable).
    \item \textbf{Associated Constraints:} A collection of rules, corresponding to the Facility-Level Constraint types detailed in. Table~\ref{tab:database_constraint_specifications}, governing its placement and relationships.
    \item \textbf{Instance Count Guideline:} A suggestion for its typical or maximum number of instances.
    \item \textbf{Semantic Tags:} Descriptive tags (e.g.,  ``Enemy,'' ``Interactable'') assisting rule application and LLM queries.
\end{itemize}

The initial population of these facilities is augmented by LLMs. We employ designed prompts that instruct the LLM to generate a specified number of facility definitions in a structured format for a given theme. For instance, a formal prompt to the LLM could be:

\textit{``Generate 10 facility definitions suitable for a survival horror 'Silent Hill hospital' environment. For each facility, provide: (a) a unique name (e.g., 'IV Stand', 'Broken Mirror'), (b) estimated dimensions (width, length, height) in meters, (c) three relevant placement constraints (refer to Table~\ref{tab:database_constraint_specifications} for Facility-Level constraint types and example formats; we omit the full information here and restrict the number of constraint to three for simplicity), and (d) 1-3 descriptive tags. The output for all facilities should be a single JSON array, where each element is an object with keys: 'name', 'dimensions', 'constraints' (an array of objects, each with 'type' and 'parameters'), and 'tags'.''}

\subsubsection{The Room Database}
\label{sec:room_database}

The Room Database contains classic room templates (e.g., ``Patient Room,'' ``Storage''  for the ``hospital'' theme), forming the building blocks for level assembly. Each template defines a space with characteristic size, function, and typical contents.

Key properties of a room template include:
\begin{itemize}
    \item \textbf{Name:} A descriptive label (e.g., ``Exam Room'').
    \item \textbf{Dimensions:} Default width, length, and height.
    \item \textbf{Characteristic Facilities:} A list of facility types (from the Facility Database) typically found within this room (e.g., a ``Storage'' room template might include ``File Cabinet,'' ``Medical Shelf,'' and ``Locker'').
    \item \textbf{Instance Frequency Guideline:} Its typical or maximum number of instances per level.
    \item \textbf{Architectural Type Definition:} A specification (e.g., ``enclosed''), a form of Room-Level Constraint (Table~\ref{tab:database_constraint_specifications}).
    \item \textbf{Associated Inter\-Room Constraints:} Optimization objective, corresponding to Room Constraints in Table~\ref{tab:database_constraint_specifications}, deciding its placement relative to other rooms or global features (e.g., ``Kitchen'' `AdjacentTo` ``Cafeteria'').
\end{itemize}

\subsubsection{The Gameplay Mechanics Database}
\label{sec:mechanics_database_construction}

The Mechanics Database, conceptualized for LLM-assisted population, contains definitions for discrete ``mechanic components,'' the ``specialized'' facilities to enable global progression (e.g., a key conceptually unlocking a door or area). Examples fitting our theme include an ``Exit Key'' or an ``Open Button.''

Key properties of a mechanic component entry, for our current simplified scope, include:
\begin{itemize}
    \item \textbf{Name:} A unique identifier for the mechanic component 
    \item \textbf{Standard Facility Constraints:} Mechanic components can have all the standard constraints applicable to regular facilities (see Facility-Level types in Table~\ref{tab:database_constraint_specifications}) to ensure their sensible placement within a room (e.g., an ``Exit Key'' might be constrained with `Near(Locker)').
    \item \textbf{Topological Constraints:} Notably, these components are governed by rules based on the topological order (\(\tau\)) of rooms to enforce logical progression. These include:
        \begin{itemize}
            \item \textit{Topological Precedence}: Ensures one component is encountered before its conceptual counterpart or consequence (e.g., an ``Exit Key'' before its ``Exit'').
            \item \textit{Topological Near/Far}: Specifies that two mechanic components should be relatively close or distant in terms of progression path (i.e., difference in topological orders of their respective rooms).
        \end{itemize}
\end{itemize}

These are detailed further under the Gameplay Mechanic Component-Level in Table~\ref{tab:database_constraint_specifications}.

Across all three database categories—Facility, Room, and Gameplay Mechanics—we parse the LLM’s structured output (e.g. JSON) into our internal representations and then subject it to expert review. Optional domain specialists (authors for now) verify each entry’s dimensions, constraint definitions, and thematic appropriateness, correcting any inconsistencies or errors introduced by the LLM. The result is a curated, theme-specific suite of databases that is both stable and internally coherent, serving as a one-time setup for rapidly assembling numerous level variations within the chosen theme.

After database curation, level instantiation begins with a designer-provided configuration. The designer defines the global dimensions of the multi‐floor environment—width \(W\), length \(L\), and floor count \(F\)—and selects which room templates to include from the Room Database, specifying target counts for each template’s `MaxInstances' limit. They also choose which gameplay mechanic components to integrate (for example, the number of “KeyFragments” for an exploration scenario). These high‐level parameters serve as input to the generation pipeline and guide the subsequent synthesis and placement phases.  

\begin{figure}[htbp]
    \centering
    \includegraphics[width=\linewidth]{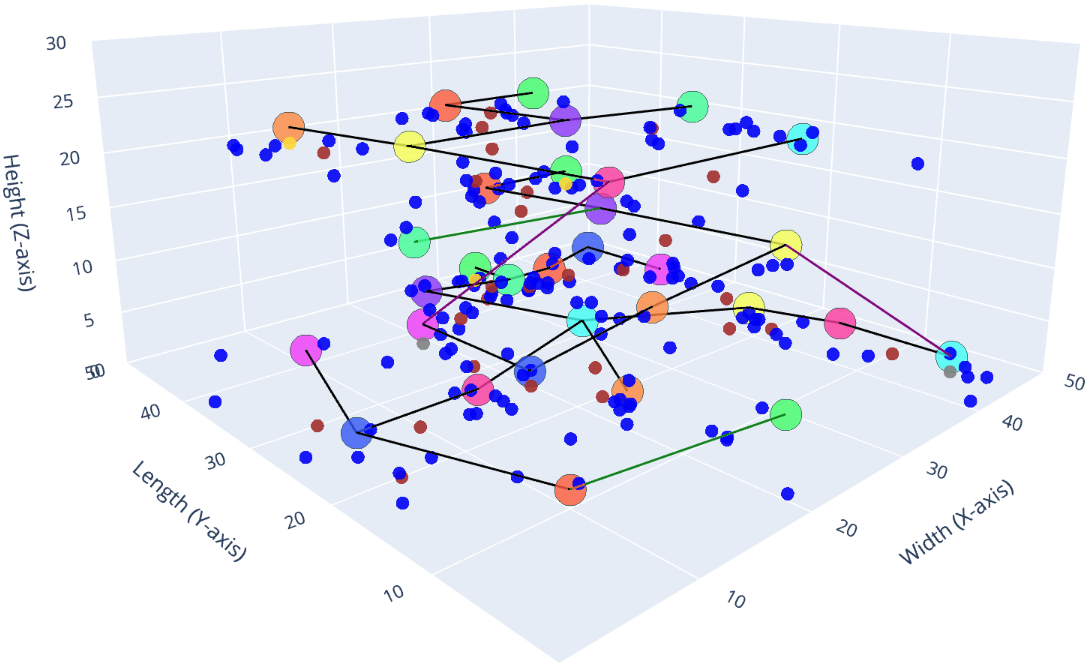}
    \caption{Generated multi-floor layout. Large colored spheres mark room centers. Small blue spheres are common facilities; brown spheres are doors, with black lines linking the two closed rooms they connect. Green lines join adjacent open rooms. Grey spheres denote stairs, with brown lines showing their vertical links.}
    \label{fig:overall_layout_sample}
\end{figure}

\begin{algorithm}[h]
\caption{Multi-Floor Room Arrangement using Room Database}
\label{alg:multi_floor_arrangement_aaai}
\begin{algorithmic}[1]
  \STATE \textbf{Input:} Level dimensions $(W,L,H)$; number of floors $F$; room database $\mathcal{D}_{\mathrm{room}}$ containing templates $\mathcal{T}$
  \STATE \textbf{Output:} Placed room instances $\mathcal{R}_{\mathrm{level}} = \bigcup_{f=0}^{F-1}\mathcal{R}_f$
  \STATE Initialize topological order $\tau \gets 1$
  \STATE Initialize placed‐room set $\mathcal{R}_{\mathrm{level}} \gets \varnothing$
  \STATE Select initial template $T_0$ (e.g.\ “Lobby”) and create $R_0$ at $(0,0,0)$
  \STATE Set $\tau(R_0)\gets\tau$, then $\tau\gets\tau+1$
  \STATE Initialize $\mathcal{R}_0\gets\{R_0\}$; add $R_0$ to $\mathcal{R}_{\mathrm{level}}$
  \STATE Initialize stack $\mathcal{S}_0\gets[R_0]$

  \FOR{$f=0$ \TO $F-1$}
    \WHILE{$\mathcal{S}_f\neq\varnothing$}
      \STATE $R_{\mathrm{curr}}\gets \textsc{Pop}(\mathcal{S}_f)$
      \FORALL{direction $d\in\{\mathrm{left,\,right,\,front,\,back}\}$}
        \STATE $R_{\mathrm{cand}}\gets \textsc{GenCandidateRoom}(R_{\mathrm{curr}},d,\mathcal{D}_{\mathrm{room}})$
        \IF{$R_{\mathrm{cand}}\neq\text{null}$ AND feasible}
          \STATE $\tau(R_{\mathrm{cand}})\gets\tau$; $\tau\gets\tau+1$
          \STATE Add $R_{\mathrm{cand}}$ to $\mathcal{R}_f$ and $\mathcal{R}_{\mathrm{level}}$
          \STATE \textsc{Push}($R_{\mathrm{cand}},\mathcal{S}_f$)
          \STATE Update usage count of its template in $\mathcal{D}_{\mathrm{room}}$
        \ENDIF
      \ENDFOR
    \ENDWHILE

    \IF{$\mathcal{R}_f\neq\varnothing$}
      \STATE $R_{\mathrm{last}}\gets\arg\max_{R\in\mathcal{R}_f}\tau(R)$
      \STATE Place “Stair” in $R_{\mathrm{last}}$ at $(x_{\mathrm{stair}},y_{\mathrm{stair}},z_{\mathrm{stair}})$
      \IF{$f < F-1$}
        \STATE Select next‐floor template $T_{\mathrm{init}}$ and create $R_{\mathrm{init}}$ on floor $f+1$ aligned at $(x_{\mathrm{stair}},y_{\mathrm{stair}})$
        \IF{$R_{\mathrm{init}}$ created successfully}
          \STATE $\tau(R_{\mathrm{init}})\gets\tau$; $\tau\gets\tau+1$
          \STATE Initialize $\mathcal{R}_{f+1}\gets\{R_{\mathrm{init}}\}$; push onto $\mathcal{S}_{f+1}$
        \ENDIF
      \ENDIF
    \ENDIF
  \ENDFOR

  \STATE \textbf{return} $\mathcal{R}_{\mathrm{level}}$
\end{algorithmic}
\end{algorithm}

\subsection{Global Room Arrangement with Greedy DFS Multi-Floor Expansion from Room Database}
\label{sec:room_arrangement}

We implement room placement as a greedy depth-first search (DFS): 
at each expansion step the candidate room with the lowest placement
penalty is selected and attached to the current frontier.
Given global dimensions \((W,L,H)\) and \(F\) floors, each room instance \(R\) inherits its template’s dimensions and is assigned a unique topological order \(\tau\) for later mechanic integration. Algorithm~\ref{alg:multi_floor_arrangement_aaai} begins by creating \(R_0\) on floor \(f=0\) from an initial template (e.g.\ “Lobby”), setting \(\tau(R_0)=1\), and pushing it onto stack \(\mathcal{S}_0\). For each floor \(f\), we pop the current room \(R_{\mathrm{curr}}\) from \(\mathcal{S}_f\) and attempt to generate candidates in the four cardinal directions by querying the Room Database for templates that physically fit, respect instance-frequency limits, and satisfy inter-room constraints (Table~\ref{tab:database_constraint_specifications}); we then select the candidate with the lowest placement penalty. Valid candidates are added to \(\mathcal{R}_f\) and \(\mathcal{R}_{\mathrm{level}}\), pushed onto \(\mathcal{S}_f\), and assigned the next \(\tau\). Once \(\mathcal{S}_f\) is empty, we place a “Stair” facility in the room with the highest \(\tau\), and if \(f<F-1\) we instantiate and align a new room \(R_{\mathrm{init}}\) on floor \(f+1\) beneath the stair, assign its \(\tau\), and seed \(\mathcal{S}_{f+1}\). The resulting multi-floor layout, showing spatial relationships and connectivity, is illustrated in Figure~\ref{fig:overall_layout_sample}.

\subsection{Local Single-Room Facility Layout Optimization with Facility Database}
\label{sec:single_room_optimization}

Each room \(R\) is first populated with its template’s fixed facilities at predetermined positions.  The remaining variable facilities \(F_R=\{f_1,\dots,f_n\}\) are given initial candidate placements (e.g.\ by random sampling within the room).

We then solve for the Facility Layout Problem (FLP):
\[
x^* \;=\;\arg\min_{x\in\mathcal{X}(R)}\;\Obj(R; x)
\]
where \(x\) encodes the positions, orientations of all \(f\in F_R\), 
\begin{equation}\label{eq:room_objective}
\Obj(R; x)
= \;\alpha\,P_{\rm local}(R; x)
\;+\;\beta\,C(R; x)
\;+\;\gamma\,S(R; x)
\end{equation}
with
\[
P_{\rm local}(R; x)
=\sum_{f\in F_R}\Bigl[p_{\rm overlap}(f)+p_{\rm cons}(f)\Bigr],
\]
\[
C(R; x)
=\sum_{i<j}\frac{1}{d(f_i,f_j)+\epsilon}, 
S(R; x)
=\max_{g\in G_R}\min_{f\in F_R}d(g,f).
\]
Here:
\begin{itemize}
  \item \(p_{\rm overlap}(f)\) penalizes collisions with facilities and walls.
  \item \(p_{\rm cons}(f)\) sums penalties for violating constraints. All constraints are implemented as soft; high penalties simulate hard constraints. 
  \item \(d(f_i,f_j)\) is the Euclidean distance between facilities.
  \item \(G_R\) is a dense grid of test points in \(R\), typically sampled with unit length, so \(S\) measures worst‐case sparsity.
  \item \(\alpha, \beta,\gamma\) weights are properly fine tuned in experiments.
\end{itemize}

We minimize \(\Obj(R;x)\) via Simulated Annealing (SA): at each iteration, randomly perturb \(x\), accept or reject based on the change in \(\Obj\), and gradually reduce temperature. The final \(x^*\) yields the optimized layout of all standard facilities in room \(R\).

\subsection{Integration of Gameplay Mechanics via Topology-Aware Optimization}
\label{sec:mechanics_integration}

After all rooms are populated with their optimized facility layouts and an overall topological order \(\tau\) is established, we integrate progression-dependent gameplay mechanics.  Each of the \(K\) mechanic components \(M=\{m_1,\dots,m_K\}\) must be assigned to one of the pre-populated rooms \(R\)'s centers.

We cast this as an optimization problem:
\begin{equation}\label{eq:assignment}
\mathbf{r}^* \;=\;\arg\min_{\mathbf{r}=(r_1,\dots,r_K)\in R^K} \;F(\mathbf{r}),
\end{equation}
where \(\mathbf{r}\) assigns each \(m_i\) to room \(r_i\), and
\begin{equation}\label{eq:fitness}
\begin{aligned}
F(\mathbf{r})
&=\,w_1\sum_{(i,j)\in\mathcal{P}}\bigl[\max\bigl(0,\;\tau(r_i)-\tau(r_j)\bigr)\bigr]^2\\
&\quad+\;w_2\sum_{i=1}^K C_{\mathrm{std}}(m_i,r_i)\\
&\quad+\;w_3\sum_{(i,j)\in\mathcal{N}}\bigl[\max\bigl(0,\,|\tau(r_i)-\tau(r_j)|-d^{\max}_{ij}\bigr)\bigr]^2\\
&\quad+\;w_4\sum_{(i,j)\in\mathcal{F}}\bigl[\max\bigl(0,\,d^{\min}_{ij}-|\tau(r_i)-\tau(r_j)|\bigr)\bigr]^2
\end{aligned}
\end{equation}

\noindent
Here:
\begin{itemize}
  \item \(\mathcal{P}\) enforces TopologicalPrecedence (\(\tau(r_i)<\tau(r_j)\));
  \item \(C_{\mathrm{std}}\) penalizes Standard Facility Constraints;
  \item \(\mathcal{N},\mathcal{F}\) enforce TopologicalNear/Far.
\end{itemize}

We employ SA to optimize the candidate assignments \(\mathbf{r}\), using the similar hyperparameter as in single-room FLP.  Once converged, each component \(m_i\) is placed in its room with greedy strategy to minimize the local intra-room constraint cost without re-optimizing the existing facility layout.

As a prototype, we integrate \textbf{key-and-lock puzzles}: e.g.\ ensure the “Exit Key” sits in room \(R_K\) with \(\tau(R_K)<\tau(R_L)\) for the corresponding “Exit Door” in \(R_L\).  By tuning the weights \(w_1\dots w_4\) and distance thresholds \(d^{\max},d^{\min}\), we generate varied key-placement strategies.  This modular framework will be extended to more complex, interdependent puzzles in future work.

\begin{table*}[ht]
\small
\centering
\begin{tabularx}{\textwidth}{@{}l l l X@{}}
\toprule
\textbf{Group}    & \textbf{Algorithm Strategy} & \textbf{Mathematical Expression} & \textbf{Implementation Approach} \\ 
\midrule
Baseline & BFS-based Balanced Layout 
 & 
$\displaystyle
\text{RoomScore}_{B}
= w_{\text{start}}\frac1{d_{\text{start}}}
+ w_{\text{end}}\frac1{d_{\text{end}}}
$
 & Compute BFS distances from all rooms to the start and end rooms;
   select the top candidate. \\[1mm]

Exploration & Random Walk Dispersion 
 & 
$\displaystyle
\text{RoomScore}_{E}
= \sigma(d_{\text{neighbors}})
\times (1 - \rho_{\text{keys}})
$
 & Use Monte Carlo simulation to identify low key-density areas and place 3 keys per floor. \\[1mm]

Speedrun & Centrality Optimization 
 & 
$\displaystyle
\text{RoomScore}_{S}
= \frac{C_{\text{closeness}}}
       {\max\bigl(1,\;\text{PathLen}_{\text{start-end}}\bigr)}
$
 & Compute all-pairs shortest paths and choose the room with highest
   closeness centrality. \\ 
\bottomrule
\end{tabularx}

\caption{Algorithm Strategies for Phase 1 Experimental Groups}
\label{tab:phase1_alg}
\end{table*}

\subsection{Two-Phase Level Repair}
\label{sec:level_repair_method}

To ensure robust navigability in the generated multi-floor levels, our framework employs a two-phase repair strategy. The first phase involves an \textbf{offline geometric correction} pass. Immediately following the full level generation, this automated step analyzes the layout, primarily using flood-fill algorithm within each room, to identify and resolve clear structural blockages (especially around doorways) by minimally repositioning problematic adaptable facilities. The second phase consists of \textbf{in-engine agent-based validation and final repair}. Here, the level is processed within a game engine environment where an autonomous agent performs systematic pathfinding tests across all rooms. Any remaining critical blockages that prevent successful traversal trigger a final targeted removal of the obstructing adaptable facilities. The specific operational parameters, success criteria, and metrics for this in-engine validation and repair process are detailed further in next section.

\section{Experiment}
\label{sec:experiment}

We evaluated our framework via a three-phase experimental pipeline: Python-based generation and repair, Unity-based repair and rerun validation, and simulation experiments integrated with Unity and commercial game engines. These experiments validate the structural playability, navigational effectiveness, and adaptability of the generated 3D levels.

\subsection{Experimental Design and Data Generation}
\label{sec:experimental_design_datagen}

To comprehensively evaluate our framework, we generated a substantial dataset of multi-floor 3D levels. 1000 unique levels (each typically 3 floors within \(50 \times 50 \times 30\) unit dimensions, using unique random seeds) were generated for \textit{each} of six distinct experimental group configurations, resulting in a total of 6000 levels.

The \textbf{core generation pipeline} included:
\begin{enumerate}[label=(\alph*)]
    \item Construction of three Databases for all architectural components and their constraints with GPT-4o
    \item Employment of the multi-floor room arrangement algorithm to assemble the level structure from selected room templates, establishing a topological order (\(\tau\))
    \item Optimization of the internal layout of standard facilities within each room instance
    \item Application of mechanics based on the experiment group
\end{enumerate}

The key variation across six groups lay in the method used for integrating gameplay mechanics, specifically to achieve distinct pacing profiles: a balanced experience (Baseline), an incentive for thorough exploration (Exploration), and an optimized path for rapid completion (Speedrun). The experimental groups were defined as follows:

\paragraph{Algorithmic Key-Placement Groups (A-Groups)}
Three groups used different specifically tuned algorithms (Table~\ref{tab:phase1_alg}) for placing key-related items, serving as benchmarks.
\begin{itemize}
    \item \textit{A-Baseline:} This strategy employed a Breadth-First Search (BFS) to find a room topologically balanced between the start and end points of a floor, using an objective \( \text{RoomScore}_{B} = w_{\text{start}} \cdot \frac{1}{d_{\text{start}}} + w_{\text{end}} \cdot \frac{1}{d_{\text{end}}} \).
    \item \textit{A-Exploration:} This approach used Monte Carlo methods constructing key density fields  \(\rho(g) = \sum_{k \in \text{Keys}} \exp(-||g - k||^2 / 2\sigma^2)\) to disperse multiple ``KeyFragment'' components in low-density areas.
    \item \textit{A-Speedrun:} This strategy utilized graph centrality measures (e.g., Closeness Centrality, \(C(R) = (N - 1) / \sum_{j \neq R} d_{Rj}\), via Floyd-Warshall) to place a ``FloorKey'' in topologically central rooms.
\end{itemize}

\paragraph{Database-Simulated Key-Placement Groups (DB-Groups)}
Three groups used the same core level architecture generation pipeline as A-Groups. However, key-placement pacing was achieved \textit{only} by configuring topological constraints of mechanic components from our Gameplay Mechanics Database (Table~\ref{tab:database_constraint_specifications}).
\begin{itemize}
    \item \textit{DB-Baseline:} A single ``FloorKey'' component was configured with a `TopologicalNear' constraint parameterized to target the midpoint of the current floor's \(\tau\) range.
    \item \textit{DB-Exploration:} Multiple ``KeyFragment'' components were instantiated with `TopologicalPrecedence' relative to a conceptual floor exit and strong `TopologicalFar' constraints between the fragments themselves.
    \item \textit{DB-Speedrun:} A single ``FloorKey'' component's placement was guided by a `TopologicalNear' constraint parameterized to target the floor's start room (lowest \(\tau\)) or a topologically central room.
\end{itemize}

\subsubsection{Analytical Approach}
\label{sec:analytical_approach}
We analyzed generated levels from two perspectives:

\begin{enumerate}
\item \textbf{Core Pipeline Performance:} Aggregated results from all generated levels were used to evaluate the framework’s overall \textbf{validity, structural stability, and efficiency}—including generation success, repair effectiveness, and architectural coherence.
\item \textbf{Key-Placement and Mechanics Adaptability:} We compared outcomes across six groups, contrasting DB-Groups (database-parameterized method) with A-Groups (algorithmic method). This analysis assessed whether parameterized mechanics are \textbf{adaptable} and \textbf{controllable} to have similar pacing as dedicated algorithmic methods.
\end{enumerate}

\begin{figure*}[ht]
    \centering
    \includegraphics[width=0.49\textwidth]{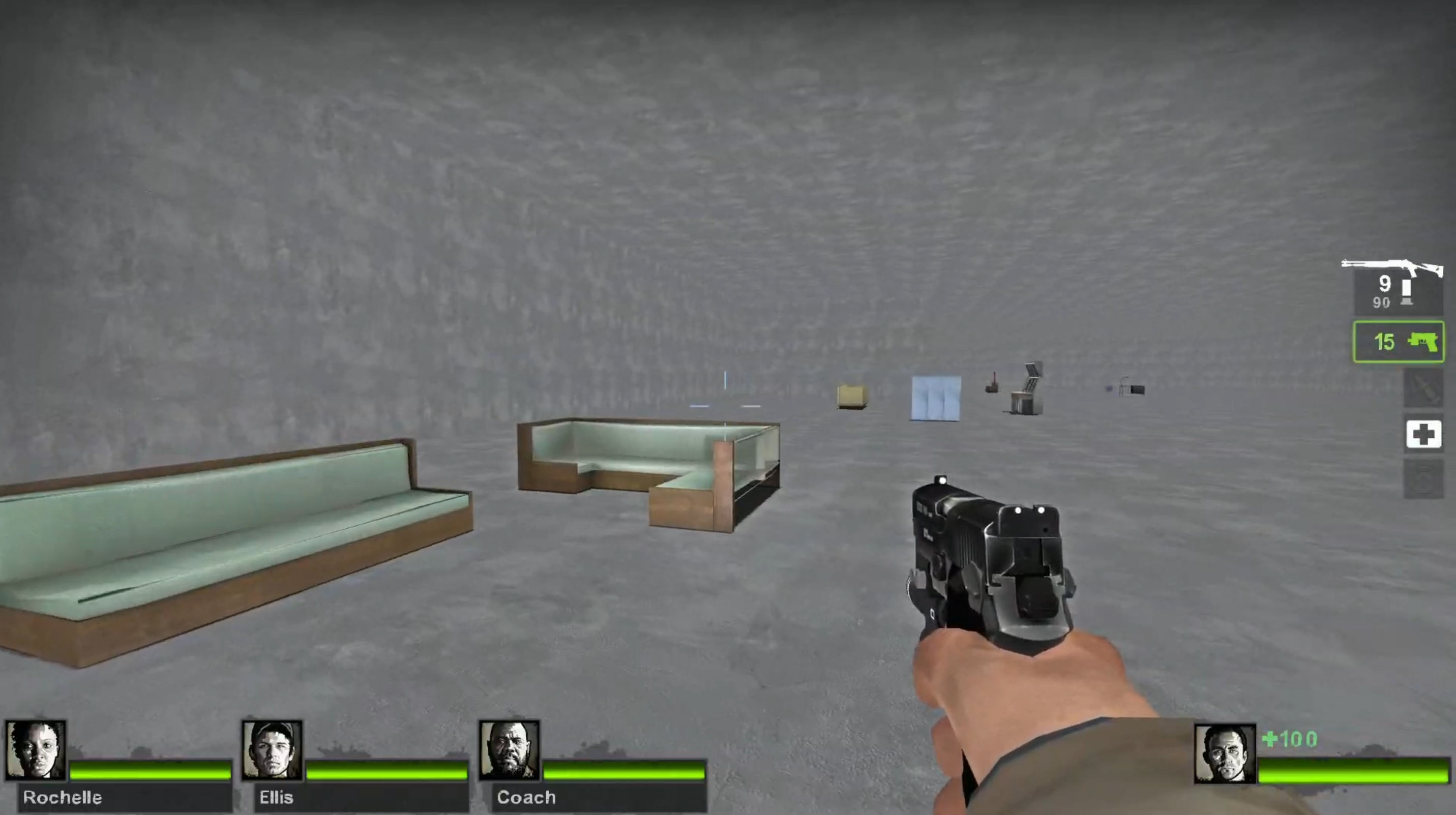}
    \hfill
    \includegraphics[width=0.49\textwidth]{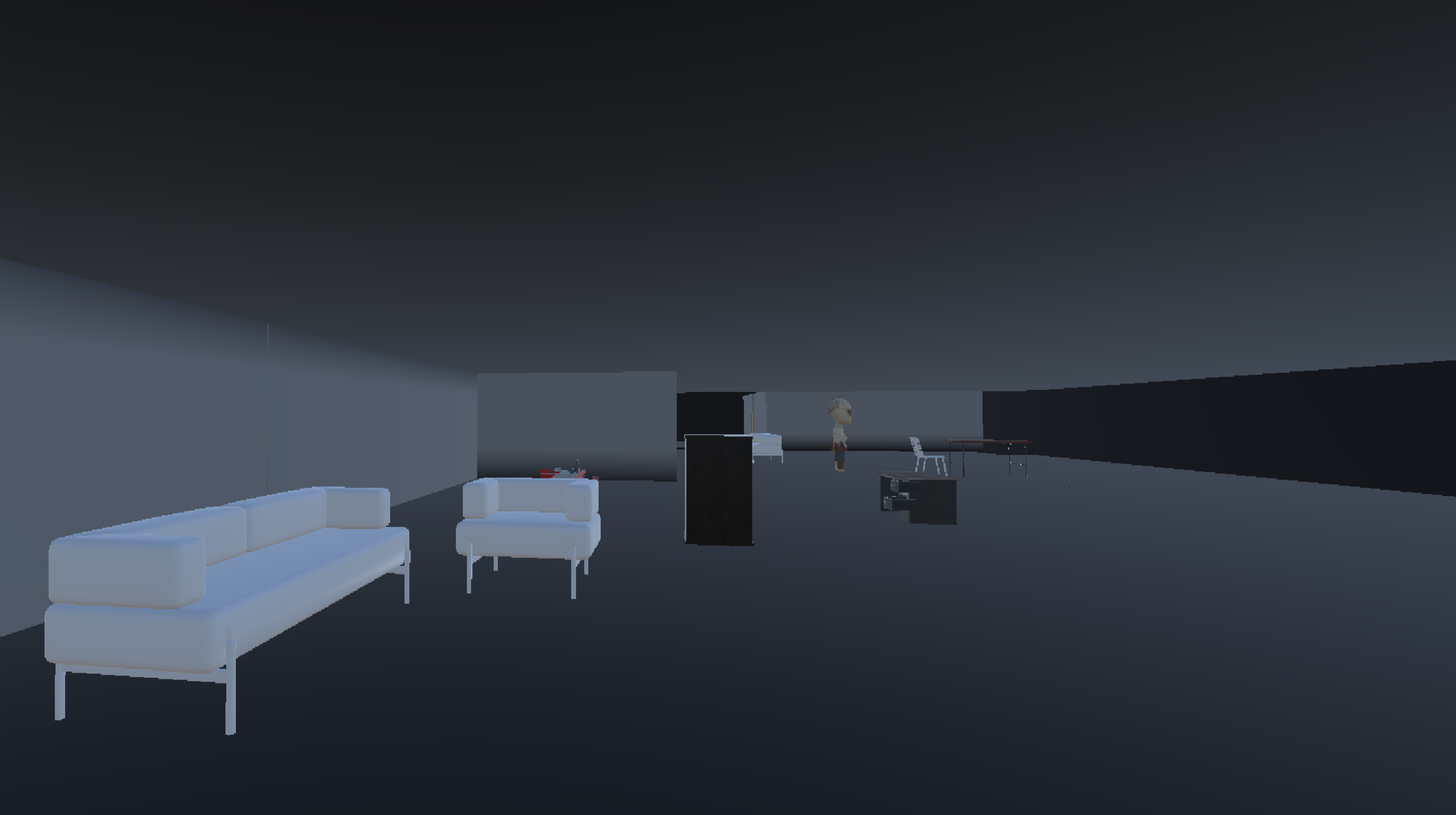}
    \caption{Illustration of the same procedurally generated layout in two different simulation environments. 
    \textbf{Left:} A screenshot from the L4D2, where the room structure and major facilities (e.g.\ couches, cabinets) have been imported into a commercial FPS game. 
    \textbf{Right:} The same spatial configuration rendered in our internal Unity simulation, where 
    an autonomous agent navigates and collects keys. }
    \label{fig:sim_comparison}
\end{figure*}

\begin{figure*}[h]
    \centering
    \includegraphics[width=\textwidth]{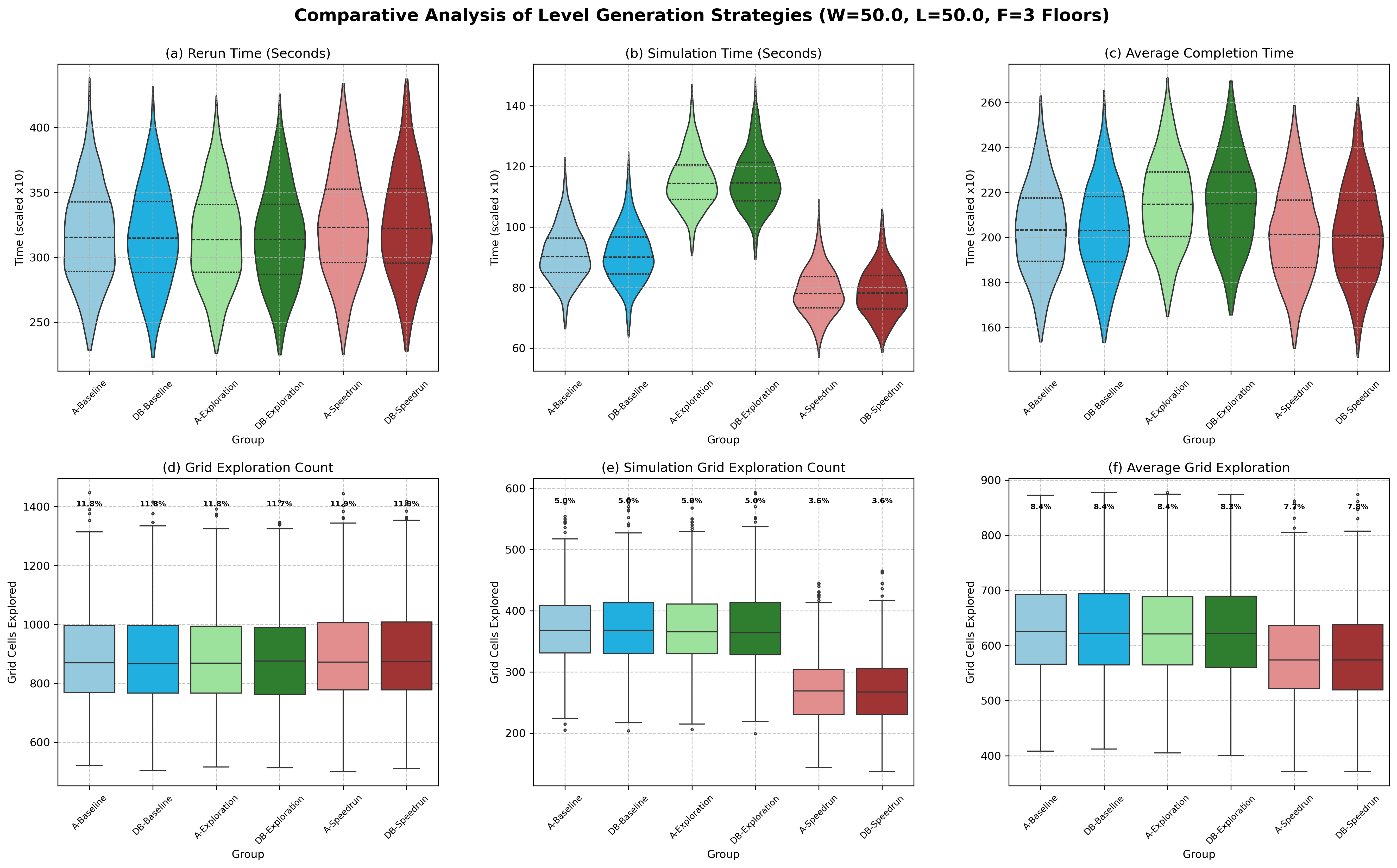} 
    \caption{Comparative analysis of level generation strategies (W=50.0, L=50.0, F=3 Floors). The top row (a-c) shows violin plots for Rerun Time, Simulation Time, and Average Completion Time. The bottom row (d-f) displays boxplots for Grid Exploration Count, Simulation Grid Exploration Count, and Average Grid Exploration. For boxplots, mean coverage ratios (\%) are annotated above each group. Data represents valid, non-abnormal levels.}
    \label{fig:six_group_analysis}
\end{figure*}

\subsection{Hyperparameters, Repair, and Simulation}
\label{sec:params_and_sim}

Our facility‐layout optimization uses SA with an initial temperature of 1000, cooling rate 0.95, and 1000 iterations. The objective combines three weighted terms: cluster degree ($\beta$=100), sparsity ($\gamma$=500), and total penalty ($\alpha$=1). Individual constraint penalties are fine tuned: axis placement $w_{\text{axis}}$=20, proximity $w_{\text{near}}$=10 ($d_{\min}$=5) and $w_{\text{far}}$=15 ($d_{\max}$=10), visibility $w_{\text{can\_see}}$=2, overlap $w_{\text{overlap}}$=30, boundary $w_{\text{bounds}}$=30, alignment $w_{\text{align}}$=15, orientation $w_{\text{orient}}$=20, and focus $w_{\text{focus}}$=10. Room‐level placement reuses $w_{\text{axis}}$ for $w_{\text{pos\_room}}$, and applies $w_{\text{near}}$/$w_{\text{far}}$ for $w_{\text{adj}}$/$w_{\text{sep}}$. Mechanic integration employs the same SA hyperparameters and adds: $w_{1}=50$, $w_{2}=1$, $w_{3}=10$ ($d^{\max}_{ij}$=4), and $w_{4}=15$ ($d^{\min}_{ij}$=3). A-Baseline sets $w_{\text{start}}=0.5$ and $w_{\text{end}}=0.5$; A-Exploration uses a Gaussian kernel with $\sigma=3$.

All in-engine repair, rerun validation, and Unity simulations employ the same agent settings: movement speed 10 u/s, angular speed 100, collision via sphere cast (radius 1 u), and NavMesh A* pathfinding. During repair, the agent visits rooms in topological order ($\tau$), with a 10 s timeout per room; any blocking facility is first repositioned and then removed if still obstructing. Levels that fail to connect all rooms within a cumulative 1000 s are marked “unrepairable.” In rerun validation, the same agent retraverses every room, logging grid cells explored and total traversal time; runs exceeding the same upper bound are flagged ``abnormal'' and excluded.

To verify our industry compatibility, generated layouts (rooms and facilities in JSON) are converted to Valve Map Format (``.vmf''): rooms instantiate predefined wall modules and facilities map to placeholder assets in \emph{Left 4 Dead 2}, with proportional scaling. The resulting \texttt{.vmf} files load into the L4D2 Hammer editor (Figure~\ref{fig:sim_comparison}, left), confirming integration. Finally, full gameplay simulations in Unity—where the agent collects keys in topological order—yield quantitative navigability metrics (traversal time, grid coverage; Figure~\ref{fig:six_group_analysis}), validating that our framework produces playable 3D levels and supports directed gameplay objectives. 

\section{Results} 
\label{sec:results}

\begin{figure}[h]
    \centering
    \includegraphics[width=\linewidth]{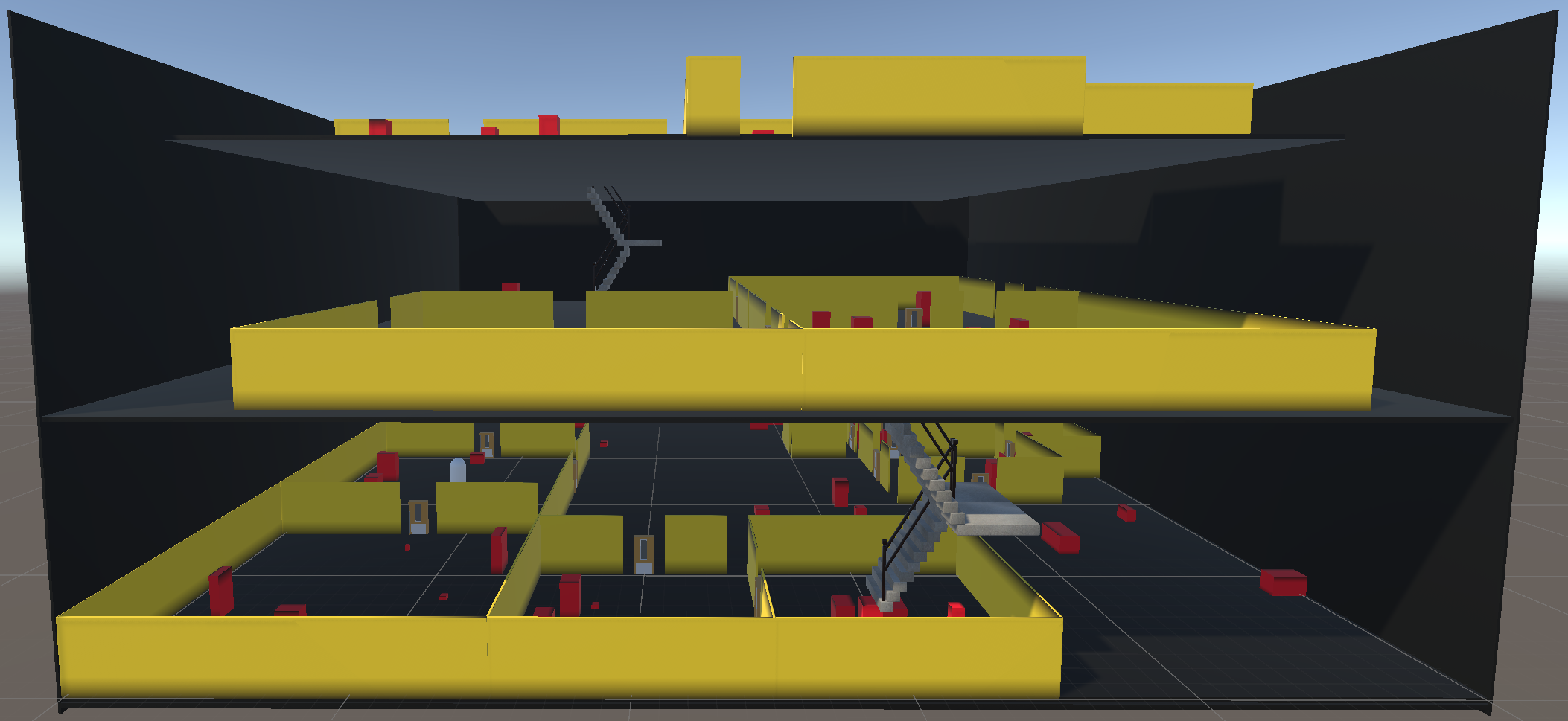} 
    \caption{Example of a multi-floor 3D level rendered in Unity for the simulation experiments. Yellow volumes define room spaces, red objects are placeholder facilities, doors and staircases provide connectivity for the navigating agent.}
    \label{fig:unity_simulation_env_example}
\end{figure}

This section presents an analysis of our generated levels. We evaluate outcomes from a total of 6000 level generation instances across six experimental groups, focusing on metrics including generation validity, repair statistics, agent-based simulation times, and grid exploration. The core of our analysis compares three key-placement strategies—Baseline, Exploration, and Speedrun—when implemented using traditional algorithmic approaches versus when simulated via parameterization of our database-driven framework.

\subsection{Overall Performance and Data Overview}
\label{subsec:data_overview}

\begin{table*}[htbp]
\centering
\resizebox{\textwidth}{!}{%
\begin{tabular}{l|cc|cc|cc} 
\toprule
\textbf{Metric} & \textbf{A-Baseline} & \textbf{DB-Baseline} & \textbf{A-Exploration} & \textbf{DB-Exploration} & \textbf{A-Speedrun} & \textbf{DB-Speedrun} \\
\midrule
Repair Time (s) & 668.18 \(\pm\) 172.93 & 667.55 \(\pm\) 173.11 & 662.72 \(\pm\) 169.04 & 661.86 \(\pm\) 170.25 & 674.91 \(\pm\) 170.46 & 675.16 \(\pm\) 171.92 \\
Facilities Removed & 9.18 \(\pm\) 4.31 & 9.18 \(\pm\) 4.33 & 9.13 \(\pm\) 4.33 & 9.13 \(\pm\) 4.34 & 9.14 \(\pm\) 4.32 & 9.15 \(\pm\) 4.33 \\
Rerun Time (s) & 316.59 \(\pm\) 37.54 & 315.95 \(\pm\) 37.94 & 315.11 \(\pm\) 37.09 & 314.97 \(\pm\) 37.96 & 325.10 \(\pm\) 40.18 & 325.33 \(\pm\) 41.27 \\
Simulation Time (s) & 91.01 \(\pm\) 8.74 & 90.95 \(\pm\) 9.23 & 115.28 \(\pm\) 8.92 & 115.40 \(\pm\) 9.45 & 78.64 \(\pm\) 7.67 & 78.62 \(\pm\) 7.94 \\
Avg. Completion Time (s) & 203.80 \(\pm\) 20.11 & 203.45 \(\pm\) 20.39 & 215.20 \(\pm\) 19.97 & 215.18 \(\pm\) 20.45 & 201.87 \(\pm\) 20.55 & 201.98 \(\pm\) 21.04 \\
Grid Exploration & 886.18 \(\pm\) 165.44 & 885.59 \(\pm\) 164.92 & 881.64 \(\pm\) 163.96 & 880.14 \(\pm\) 165.03 & 891.50 \(\pm\) 168.84 & 893.62 \(\pm\) 172.39 \\
(Coverage \%) & (11.8\%) & (11.8\%) & (11.8\%) & (11.7\%) & (11.9\%) & (11.9\%) \\
Sim. Grid Exploration & 372.92 \(\pm\) 59.68 & 373.23 \(\pm\) 61.31 & 372.29 \(\pm\) 60.86 & 371.60 \(\pm\) 62.09 & 269.82 \(\pm\) 54.72 & 269.67 \(\pm\) 55.18 \\
(Sim. Coverage \%) & (5.0\%) & (5.0\%) & (5.0\%) & (5.0\%) & (3.6\%) & (3.6\%) \\
Avg. Grid Exploration & 629.55 \(\pm\) 92.32 & 629.41 \(\pm\) 92.40 & 626.97 \(\pm\) 92.33 & 625.87 \(\pm\) 93.18 & 580.66 \(\pm\) 86.67 & 581.64 \(\pm\) 88.06 \\
(Avg. Coverage \%) & (8.4\%) & (8.4\%) & (8.4\%) & (8.3\%) & (7.7\%) & (7.8\%) \\
\bottomrule
\end{tabular}
}
\caption{Key Metrics for Valid \& Non-Abnormal Levels Across Six Experimental Groups. ``A-'' prefixes denote traditional algorithmic implementations; ``DB-'' prefixes denote simulations using our database-driven approach. Grid exploration metrics include average coverage ratio (\%).}
\label{tab:six_group_metrics}
\end{table*}

Across 6000 generation attempts, our pipeline confirmed robust performance: 5728 levels (95.47\%) were validated as successfully repaired and free of subsequent anomalies. ``Unrepairable'' levels—where the repair process could not restore navigability—accounted for 208 cases (3.47\%). An additional 64 levels (1.07\%) passed repair but were flagged as ``abnormal'' during rerun validation, typically due to extreme traversal metrics reflecting rare interactions between level geometry and agent pathfinding. Most groups exhibited a low number of abnormal levels (A-Baseline: 10, DB-Baseline: 10, A-Speedrun: 12, DB-Speedrun: 9), with only A-Exploration (19) and DB-Exploration (16) slightly higher due to multi-key complexity. Irreparable levels usually resulted from highly constrained initial placements or dense multi-floor configurations; abnormal levels arose from rare seeds creating atypically complex paths.

Multiple metrics in Table~\ref{tab:six_group_metrics} indicate the baseline quality of generated layouts. Average repair times (approx. 660--675s, resulting from agent‑based trial‑and‑error loop in Unity environment) and the consistently low average number of facilities removed (about 9 per level across all groups) suggest that the pipeline produces coherent structures requiring little correction. This indicates that room placements and intra-room layouts are generally robust and require only minor adjustment for full agent traversability under strict constraints.

Agent exploration metrics further support the richness of the generated spaces. For valid levels, the average ``Grid Exploration Count'' (Fig.~\ref{fig:six_group_analysis}d) shows agents traverse a substantial area (coverage ratios of 11.7--11.9\%), and the ``Simulation Grid Exploration Count'' (Fig.~\ref{fig:six_group_analysis}e) is consistently lower, implying that objective-based navigation remains meaningful. Moreover, the fact that these exploration patterns and completion times (Fig.~\ref{fig:six_group_analysis}b,c) can be modulated by key-placement strategies (see below) demonstrates that the generated level structures can support distinct gameplay pacings, reflecting structural quality beyond mere traversability.

\subsection{Analysis of Key-Placement Strategies and Database Adaptability}
\label{subsec:comparative_analysis_strategies}

This analysis focuses on the valid and non-abnormal levels from each of the six experimental groups. \textbf{Our primary objectives} are to: (1) validate that our database-driven parameterization of gameplay mechanics can influence game flow and produce varied gameplay experiences, tuned to different playstyles, even within similarly structured levels; and (2) show that this database-driven approach can effectively simulate the outcomes of distinct, traditional algorithmic key-placement strategies through simpler configuration. 

Figure~\ref{fig:unity_simulation_env_example} shows a typical multi-floor 3D environment generated by our system and rendered in Unity, representative of the architectural complexity within which our agent simulations were conducted. Yellow volumes define room spaces, red objects are placeholder facilities, white capsule is the agent, and staircases and doors provide connectivity. The quantitative results of agent performance within such environments are presented in Table~\ref{tab:six_group_metrics} and visualized in Figure~\ref{fig:six_group_analysis}.

\paragraph{Impact of Gameplay‐Mechanic Placement on Flow and Exploration}

Figure~\ref{fig:six_group_analysis} and Table~\ref{tab:six_group_metrics} show that altering the topological rules in the Mechanics Database produces clear, quantifiable shifts in agent behaviour.  

\textbf{Time-based metrics:} ``Simulation Time''—the minimal time to complete the level on an optimal path—differs markedly across groups: \textit{DB-Speedrun} $78.62\pm7.94$ s, \textit{DB-Baseline} $90.95\pm9.23$ s, and \textit{DB-Exploration} $115.40\pm9.45$ s. As intended, keys forced \emph{near} the start (Speedrun) are found quickest, whereas \emph{far} placement (Exploration) yields the slowest, most variable searches. Full traversal (``Rerun Time'') remains similar (\textit{DB-Baseline} 315.95 s, \textit{DB-Exploration} 314.97 s, \textit{DB-Speedrun} 325.33 s) because the room graph is unchanged; these search-time shifts drive corresponding changes in ``Average Completion Time'' (201.98 s, 203.45 s, and 215.18 s, respectively), confirming that progression pace is controllable via simple database parameters.

\textbf{Grid-based metrics:} Search efficiency is echoed in space coverage. For each entire level, the agent visits only $269.67$ cells (3.6 \%) in \textit{DB-Speedrun}, indicating partial validity, but $373.23$ (5.0 \%) and $371.60$ (5.0 \%) in \textit{DB-Baseline} and \textit{DB-Exploration}. Despite the longer search time, Exploration’s covered area is not much larger, suggesting its \textit{TopologicalFar} rule reasonably spaces objectives along the main route rather than hiding them in peripheral niches. Across a full run, average coverage settles at 7.8 \% (\textit{Speedrun}) versus 8.4 \% (\textit{Baseline}) and 8.3 \% (\textit{Exploration}). Hence, purely by changing topological distance constraints in the Mechanics Database, our framework produces distinct, predictable profiles for both temporal pacing and spatial exploration.

\paragraph{Database-Driven Simulation of Algorithmic Pacing Strategies}

A central aim was to prove that our Gameplay Mechanics Database can, through parameterization, effectively simulate outcomes typically requiring distinct, complex algorithms. Table~\ref{tab:six_group_metrics} and Figure~\ref{fig:six_group_analysis} provide strong evidence for this.
For the Baseline strategy, DB-Baseline (Avg. Completion Time: 203.45s \(\pm\) 20.39s) closely mirrors A-Baseline (203.80s \(\pm\) 20.11s). Their 95\% CIs for this metric ([202.15, 204.74] vs. [202.52, 205.08]) are highly overlapping, and other metrics like ``Simulation Time'' (90.95s vs. 91.01s) and ``Sim. Grid Exploration'' (373.23 vs. 372.92) are nearly identical.
For the Exploration strategy, DB-Exploration (Avg. Completion Time: 215.18s \(\pm\) 20.45s) aligns remarkably well with A-Exploration (215.20s \(\pm\) 19.97s), with virtually identical 95\% CIs ([213.88, 216.49] vs. [213.93, 216.47]). Their ``Simulation Times'' are also extremely close (115.40s vs. 115.28s).
Similarly, for the Speedrun strategy, DB-Speedrun (Avg. Completion Time: 201.98s \(\pm\) 21.04s) effectively matches A-Speedrun (201.87s \(\pm\) 20.55s), with overlapping CIs and almost identical ``Simulation Times'' (78.62s vs. 78.64s) and ``Sim. Grid Exploration'' counts (269.67 vs. 269.82).

These consistencies across all three pacing strategies strongly indicate that by configuring topological constraints within our Mechanics Database, we can achieve targeted gameplay flow outcomes that are quantitatively same to those produced by more complex, specialized algorithms.
This highlights the adaptability and efficiency of our approach: diverse gameplay experiences can be authored through high-level parameterization of database components rather than disparate algorithmic implementations.

\section{Conclusion}
\label{sec:conclusion}

This paper introduced a modular framework for generating multi-floor 3D game levels, distinguished by its novel use of Large Language Models for the \textit{offline} creation of \textbf{themed, reusable, controllable, and richly constrained databases for facilities, rooms, and mechanic components}. Our multi-phase pipeline leverages these curated databases, enabling designers to select desired architectural and mechanical elements and, through straightforward parameterization, assemble complex, playable 3D environments via constraint-based optimization at both inter-room and intra-room scales, followed by a robust repair process.

Our primary experimental objective was to verify the viability and benefits of this database-centric paradigm, shifting away from mostly real-time LLM interactions for level generation to a more structured, designer-guided assembly of LLM-augmented, pre-defined assets, pushing the design / plan phase ahead. The results confirmed the high stability and success rate of our core architectural generation pipeline. Furthermore, by configuring topological constraints within the Gameplay Mechanics Database, we successfully simulated diverse gameplay pacing strategies, achieving outcomes quantitatively comparable to those produced by distinct, complex algorithms. This underscores the adaptability of our modular framework, where each database can be independently extended or refined with new constraints and content to suit varied design requirements for facility interactions, room relationships, and mechanic interdependencies. We leave the analysis of user-driven workflow controls and player experience evaluations for future work.

\paragraph{Limitations} While LLMs accelerate the drafting of database content, achieving high-quality, thematically coherent, and logically sound entries currently requires significant prompt engineering and potential manual curation. Secondly, although infrequent, the generation of some unplayable levels indicates that our current constraint-solving and repair heuristics have imperfections that need further refinement. Our current implementation of gameplay mechanics via topological order, while effectively indicating adaptable pacing with key-lock puzzles, represents a simplified prototype; a more comprehensive exploration of complex, interdependent puzzle mechanics is necessary. Finally, our current validation of level effectiveness, controllability, and adaptability relies on agent-based simulations; incorporating human player studies would be crucial to assess experiential qualities such as engagement and perceived level quality.

\paragraph{Future work}  We aim to develop better prompt engineering techniques and automated validation methods to reduce the manual effort in database curation. Refining the repair heuristics and potentially integrating more advanced global constraint satisfaction methods will target the reduction of unplayable instances. A major focus will be on expanding the Gameplay Mechanics Database and its integration logic to support a richer variety of interactive puzzle systems. Crucially, we plan to conduct user studies to evaluate workflow control and player experience in the generated levels.

In conclusion, by shifting the role of LLMs towards the creation of structured, offline, reusable databases, this research offers a novel, modular, and adaptable framework for the efficient generation of complex, playable 3D game levels, offering designers greater control and flexibility.

\bibliography{references}

\end{document}